\documentclass[letterpaper]{article} 
\usepackage[]{aaai25}  
\usepackage{times}  
\usepackage{helvet}  
\usepackage{courier}  
\usepackage[hyphens]{url}  
\usepackage{graphicx} 
\usepackage{natbib}  
\usepackage{caption} 
\usepackage{algorithm}
\usepackage{algorithmic}
\usepackage{graphicx}
\usepackage{multirow}
\usepackage{amsmath}
\usepackage{amsfonts,amssymb}
\usepackage{subcaption}
\usepackage{array}
\usepackage{arydshln}
\usepackage{makecell}
\usepackage{epsfig}
\usepackage{epstopdf} 
\usepackage{newfloat}
\usepackage{listings}
\DeclareCaptionStyle{ruled}{labelfont=normalfont,labelsep=colon,strut=off} 
\floatstyle{ruled}
\newfloat{listing}{tb}{lst}{}
\floatname{listing}{Listing}
\title{Preference-Oriented Supervised Fine-Tuning: Favoring Target Model Over Aligned Large Language Models}

\author{
    Yuchen Fan, Yuzhong Hong, Qiushi Wang, Junwei Bao\thanks{Corresponding author}, Hongfei Jiang, Yang Song 
}

\affiliations{
Zuoyebang Education Technology (Beijing) Co., Ltd\\
\{fanyuchen02, hongyuzhong, wangqiushi02, jianghongfei, songyang\}@zuoyebang.com \\
baojunwei001@gmail.com
}

\begin{document}

\maketitle

\begin{abstract}
Alignment, endowing a pre-trained Large language model (LLM) with the ability to follow instructions, is crucial for its real-world applications. 
Conventional supervised fine-tuning (SFT) methods formalize it as causal language modeling typically with a cross-entropy objective, requiring a large amount of high-quality instruction-response pairs.
However, the quality of widely used SFT datasets can not be guaranteed due to the high cost and intensive labor for the creation and maintenance in practice.
To overcome the limitations associated with the quality of SFT datasets, 
we introduce a novel \textbf{p}reference-\textbf{o}riented supervised \textbf{f}ine-\textbf{t}uning approach, namely PoFT. 
The intuition is to boost SFT by imposing a particular preference: \textit{favoring the target model over aligned LLMs on the same SFT data.}
This preference encourages the target model to predict a higher likelihood than that predicted by the aligned LLMs, incorporating assessment information on data quality  (i.e., predicted likelihood by the aligned LLMs) into the training process. 
Extensive experiments are conducted, and the results validate the effectiveness of the proposed method.
PoFT achieves stable and consistent improvements over the SFT baselines across different training datasets and base models.
Moreover, we prove that PoFT can be integrated with existing SFT data filtering methods to achieve better performance, and further improved by following preference optimization procedures, such as DPO.

\end{abstract}

%
\begin{links}
    \link{Code}{https://github.com/Savannah120/alignment-handbook-PoFT/}\
\end{links}

\section{Introduction}

\begin{figure}[ht]
    \centering
    \scalebox{0.53}{
    \includegraphics[]{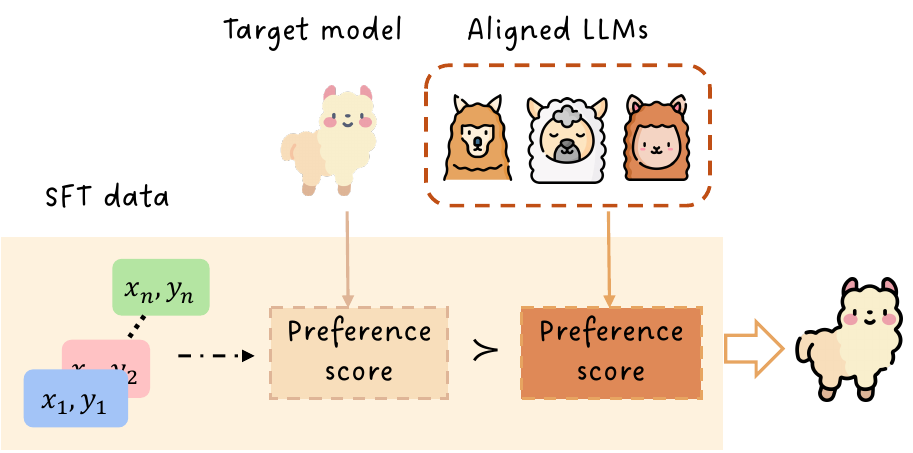}}
    \caption{The overall modeling framework of PoFT. By leveraging the Bradley-Terry ranking objective, we impose a particular preference that favors the target model over the aligned LLMs on the same SFT data. Note that the preference score is generated based on the corresponding predicted likelihood.} 
    \label{modeling structure}
\end{figure}

Large language models(LLMs) such as ChatGPT \cite{openai} have exhibited successful and potent applications in comprehending human queries and delivering plausible responses. 
This ability has proven to be crucial in real-world applications, e.g. AI assistants and recommendation systems.
To equip LLMs with this ability, the alignment methods are usually applied to pre-trained language models.
Alignment enables pre-trained models to comprehend the context and generate responses suitable to human interactions. 
Typical alignment methods can be broadly categorized into two types: Supervised Fine-Tuning (SFT) and Preference Alignment (PA).



Supervised fine-tuning (SFT) is an essential phase of alignment, wherein the task is framed as causal language modeling performed on a pre-trained language model with instruction-response data $\mathcal{D}=\{\langle x,y\rangle\}$. Generally, it leverages the cross-entropy objective function in optimization, equipping the pre-trained language model with the ability to follow instructions and generate coherent sequences. Several studies \citep{pet,adapter,hint} are dedicated to exploring SFT training strategies to enhance the alignment of LLMs. 
However, due to the intrinsic traits of modeling, the optimization process heavily depends on the availability of high-quality $\langle x,y\rangle$ data, which hinders its performance.
Traditionally, the prevalent large-scale SFT datasets in earlier research, such as Alpaca \cite{alpaca} and ShareGPT \cite{ShareGPT}, were mainly developed via AI distillation or human-and-AI interaction. Assuring the quality of these datasets can be challenging, as the filtration and curation processes demand significant human resources and efforts.

Instead of solely aligning the instruction and responses, preference alignment (PA), such as InstructGPT \cite{chatgpt} and Direct Preference Optimization (DPO) \cite{dpo}, optimizes the LLMs based on chosen-rejected data $\langle x,y^+,y^-\rangle$. These PA methods provide exceptional benefits in model alignment, enabling LLMs to align more accurately with AI/human preferences. In particular, DPO employs the Bradley-Terry (BT) ranking objective \cite{Bradley_Terry} in its optimization process to perform direct preference comparison.



Given the limitations of SFT in processing quality-limited data, we leverage the benefits of the BT preference model and incorporate it into the SFT framework, by proposing a \textbf{P}reference-\textbf{o}riented supervised \textbf{F}ine-\textbf{T}uning method, called \textbf{PoFT}.
Specifically, it applies the BT objective to different models by imposing a particular preference: favoring the target model over the aligned LLMs, given the same $\langle x,y\rangle$ data. 
Within this framework, the aligned LLMs act as baselines for the target model, prompting it to attain higher preference scores than that of the aligned LLMs on SFT data. 
Here, we assume these LLMs could discern data that contribute positively to model optimization, thereby providing valid data quality assessments.
Moreover, we would like to emphasize that we use BT model \emph{to rank models rather than to rank data}. This means we are fundamentally not a PA approach but rather an SFT approach since we require only $\langle x, y\rangle$ and not $\langle x, y^+, y^-\rangle$.
For that matter, we show our approach is indeed \emph{orthogonal} to PA since PoFT can be combined with PA methods to further enhance the overall alignment performance (e.g., first PoFT and then DPO).


Despite leveraging the preference modeling with BT, at its essence, PoFT remains faithful to the SFT paradigm, relying on instruction-response data.
As an enhanced SFT method, PoFT's objective offers a remarkable advantage over the conventional SFT objective cross-entropy (CE), i.e., PoFT is more stable and robust when training with quality-limited data. 
Specifically, the introduction of aligned LLMs provides quality assessments on each sample $\langle x,y\rangle$, which decreases its sensitivity towards the data quality.
In practice, by analyzing the gradient updates, we observe that PoFT assigns dynamic weights (namely coefficient defined in section \ref{sec:method}) to different samples $\{\langle x,y\rangle\}$ by the aligned LLMs. 
These weights guide parameter optimization, reducing the negative effect of low-quality data. 
In contrast, the CE objective treats all the data equally, without differentiating data samples based on their quality, thus exposing it to vulnerabilities to low-quality data.

In summary, our contributions are three-fold:
\begin{itemize}
    \item \textbf{Innovative SFT Training Methodology With Preference Modeling.} We present a novel method, called PoFT. This new methodology effortlessly integrates aligned LLMs for preference modeling - a fresh perspective that leads to a boost in the optimization process. 
    \item \textbf{Analytical Insight into PoFT's Stability.} Through rigorous mathematical analysis, we provide theoretical explanations that shed light on the inherent characteristics of PoFT in gradient update. 
    \item \textbf{Comprehensive Validation of Methodology.} We validate the effectiveness of PoFT through extensive experiments on different base models, demonstrating that PoFT achieves superior performance over the CE objective across diverse training datasets. Our ablation studies indicate PoFT's stability over increasing epochs and enhanced resilience to noise data. Impressively, our experiments prove that the integration of the PoFT and SFT filtering methods can lead to further performance enhancement. Moreover, the two-step training followed by DPO also shows promising alignment performance. 
    
\end{itemize}

\section{Related Work}

\subsection{Supervised Fine-Tuning}

Enabling pre-trained language models to follow human instructions, supervised fine-tuning (SFT) is a way to align LLMs' behavior with human desirability, by training on instruction-response data in a supervised fashion.

\textbf{Dataset Construction} Efforts have been made to construct diverse and complex training data, such as Orca \cite{orca} and WizardLM \cite{wizardlm}. \citet{Self-Instruct} proposed a self-improvement pipeline, which enhances LLMs by using its own generations as a bootstrap. Rather than based on human-provided instructions, \citet{backtranslation} reversely constructed instructions from the web corpus via a back-translation model. 

\textbf{Data Filtering} In addition to enhancing data complexity, some studies focus on data filtering to improve training efficiency \citep{alpagsus, instag, makesgooddataalignment, MoDS}. \citet{instag} trained a tagger based on semantics and intentions and regarded the number of tags as a complexity indicator for filtering. IFD, proposed by \citet{IFD}, is a complexity metric that identifies the discrepancies between responses and the model's generation capability. \citet{makesgooddataalignment} trained a scorer via ChatGPT to assess the complexity and quality of the data, thereby selecting ``good" data. \citet{reflection-tuning} and \citet{superfiltering} leveraged a student model to select data for training a teacher model based on the IFD scores.

\textbf{FT strategies} Multiple works have explored efficient fine-tuning strategies to enhance the alignment process \cite{pet,adapter,lora,hint}. \citet{pet} converted the provided input into cloze-style statements, thereby facilitating language models to understand the tasks. \citet{hint} transformed the instructions and examples of a task into parameter-efficient modules through an extra text encoder. Different from these strategies, PoFT proposes a training objective by modeling preference between the target model and aligned LLMs, providing a fresh perspective to enhance the optimization process.

\subsection{Preference Alignment}
By aligning training objectives with human/AI preferences, RLHF/RLAIF are particularly useful in applications that require nuanced and context-aware decisions \cite{chatgpt, openai, Anthropic}. A prominent preference alignment approach is Direct Preference Optimization (DPO) \cite{dpo}, which leverages Bradley-Terry (BT) ranking objective \cite{Bradley_Terry} to better prioritize actions based on perceived desirability. In general, the BT model estimates the probability of one item $i$ being chosen over another $j$ in a pairwise comparison, where the items are quantified with strength or quality parameters, denoted as $\lambda_i$ and $\lambda_j$ respectively, resulting in:
\begin{equation}
    \mathcal{P}(i \succ j) = \frac{\lambda_i}{\lambda_i + \lambda_j}.
\end{equation}
As for DPO, it applies the BT objective to express preferences of the policy model for the chosen-rejected pairs $\langle x,y^+,y^-\rangle$ via their expected rewards. Therefore, the preference distribution can be written as:

\begin{equation}
\begin{split}
    \mathcal{P}\left( y^+ \succ  y^- \mid x \right) &  = \sigma((r\left(x, y^+\right) - (r\left(x, y^-\right))\\
                                          &  = \frac{\exp \left(r\left(x, y^+\right)\right)}{\exp \left(r\left(x, y^+\right)\right)+ \exp\left(r(x,y^-)\right)},
\end{split}
\end{equation}
where $r\left(x, y\right)$ is a closed-form reward expression with the optimal policy in DPO's context. Subsequently, more methods are proposed to improve the preference optimization process \citep{rrhf,raft,pro,spin}.


\section{Methodology}\label{sec:method}

\subsection{Preliminary}
Typically, the cross-entropy(CE) objective for SFT training only minimizes the difference between predicted and true distributions, represented as
\begin{equation}
    L_{\text{CE}}=-\frac{1}{T_0(y)}\log p_\theta(y|x),
\end{equation}
where $T_0(y)$ refers to the length of $y$ tokenized by the target model $\theta$. Its gradient is shown in Eq. \ref{ce_grad}. 
\begin{equation}
    \begin{split}
    \nabla_\theta L_{\text{CE}} &= - \frac{1}{T_0(y)} \frac{1}{p_\theta\left(y \mid x \right)} \nabla  p_\theta\left(y \mid x \right)  \\
    \end{split}
    \label{ce_grad}
\end{equation}

\subsection{PoFT: Preference-oriented Supervised Fine-Tuning}
In this section, we introduce a novel preference-oriented fine-tuning objective that applies the Bradely-Terry model to perform preference modeling between the target model and aligned LLMs, namely PoFT. 
Given data $\{x,y\} \sim \mathcal{D}_{SFT}$, it imposes a particular preference by prioritizing the target model over the aligned LLMs. 
Accordingly, the aligned LLMS acts as a reference point guiding the target model to generate higher preference scores.
The preference score is generated based on the predicted likelihood, thus the one from aligned LLMs can be regarded as an indicator for estimating the data quality.
Assigning such a preference could diversify the effects of the SFT data, emphasizing more on high-quality data in the optimization process.
Note that the preferences are supposed to be generated by some reward model $r^*(x,y)$. Consequently, by applying the BT model, the preference distributions $\mathcal{P}(\cdot)$ can be defined as:
\begin{equation}
\begin{split}
    \mathcal{P}\left( r^*(x,y) \succ  r_{\text{LLMs}}(x,y) \mid x,y \right) \\ = \frac{\exp \left(r^*\left(x, y\right)\right)}{\exp \left(r^*\left(x, y\right)\right)+ \exp\left(r_{\text{LLMs}}(x,y)\right)},\\
     r_{\text{LLMs}}(x,y)  =  \mathbb{E}_{\text{LLM}\sim \mathcal{D}_{\text{LLMs}}}\left[r_{\text{LLM}}(x,y)\right].
\end{split}
\end{equation}
When accessing a static SFT dataset, a number of aligned LLMs (denoted as $\text{LLM}_j\in \mathcal{D}_{\text{LLM}}$,$\quad |\mathcal{D}| = M$), and a parameterized reward model $r_\theta(x,y)$ for $r^*(x,y)$, the training objective can be transformed into a binary classification problem via maximum likelihood:


\begin{small}
\begin{equation}
\begin{split}
& L_{\text{PoFT}}(\boldsymbol{\theta}) \\ 
& = -\mathbb{E}_{(x,y)\sim \mathcal{D}_{\text{SFT}}, r_\theta(x,y) \succ  r_{\text{LLMs}}(x,y)  \sim \mathcal{P}(\cdot)} \left[\log \mathcal{P}_\theta(\cdot)\right]\\
& \approx -\mathbb{E}_{(x,y) \sim \mathcal{D}_{\text{SFT}}} \left[\log \frac{\exp \left(r_\theta(x,y)\right)}{\exp \left(r_\theta(x,y)\right)+ \exp\left(r_{\text{LLMs}}(x,y)\right)}\right]\\
& =  -\mathbb{E}_{(x,y) \sim \mathcal{D}_{\text{SFT}}} \left[\log \sigma  \left(\frac{1}{M}\sum_{j=1}^M  \left( r_\theta\left(x,y\right) -r_j\left(x,y\right)\right)\right)\right],
\label{eq1}
\end{split}
\end{equation}
\end{small}
where we first impose a particular preference $\mathcal{P}\longrightarrow 1$ (hence the $\approx$) and then parameterize the BT model (i.e., $\mathcal{P}_{\theta}(\cdot)$) using rewards defined as follows: 
\begin{align*}
     r_\theta(x,y) &= \frac{1}{T_0(y)}\log p_\theta(y \mid x),\\
     r_j(x,y) &= \frac{1}{T_j(y)}\log p_j(y \mid x).
\end{align*}



In our context, we leverage \textit{the logarithm of predicted likelihood $p(y|x)$ with length normalization} as the reward function to generate \textit{preference scores}.
Specifically, the logarithm of the predicted likelihood for the target model $\log p_\theta(y|x)$ and the $j$-th aligned LLM  $\log p_j(y|x)$ are normalized by the corresponding length of the tokenized $y$, i.e., ${T}_0(y)$ and ${T}_j(y)$ respectively.
This preference score measures how likely a model would generate the response $y$ when given $x$ at a token level.
Applying the length normalization effectively addresses issues related to tokenization mismatches. Moreover, as demonstrated in \cite{simpo}, length normalization can also mitigate the impact of sequence length on the reward.

\begin{equation}
    \begin{split}
    \nabla_\theta L_{\text{PoFT}} 
    &= - \frac{1}{{T}_0(y)} \frac{1}{p_\theta\left(y \mid x \right)} \tau  \nabla  p_\theta\left(y \mid x \right),  \\
    \end{split}
    \label{PoFT_grad}
\end{equation}
where
\begin{equation}
    \tau =  \frac{\left( \prod \limits_{j=1}^M p_j(y \mid x)^\frac{1}{{T}_j(y)} \right)^\frac{1}{M}}{\left( \prod \limits_{j=1}^M p_j(y \mid x)^\frac{1}{{T}_j(y)} \right)^\frac{1}{M}+p_\theta(y \mid x)^{\frac{1}{{T}_0(y)}}}.\\
    \label{coefficient}
\end{equation}
To delve deeper into the behavior of PoFT during optimization, we examine and present the gradients for CE and PoFT loss, shown in Eq.\ref{ce_grad} and Eq.\ref{PoFT_grad}, respectively. By comparison, it can be observed that PoFT's gradient contains an extra coefficient, which is outlined in Eq. \ref{coefficient}.
This coefficient indicates that the gradient is positively related to $p_j(y|x)$, which indicates the assessment of $\langle x,y\rangle$ from the aligned LLMs. Intuitively, it allows for a more nuanced and dynamic optimization process, accounting for the unbalanced quality of the SFT datasets. Instead of assigning equal weights to all data, PoFT utilizes the aligned LLMs to direct optimization by diversifying the impacts of different samples on the gradient update. Accordingly, PoFT is proficient in alleviating the influence of lower-quality data, concentrating focus on data with a higher preference score. Thus, PoFT demonstrates its stability for the quality-limited data, compared to the conventional SFT methods.


\section{Experiment}

\begin{table*}[ht]
  \centering
  \small
  \setlength{\extrarowheight}{3pt} 
  \begin{tabular}{ll|l|cccccc|llc}
    \Xhline{1px}
    \textbf{Base} & \textbf{FT} &\textbf{Datasets}  & \textbf{Arc} & \textbf{Truthful.} & \textbf{Wino.} & \textbf{GSM8k} & \textbf{HellaS.}  & \textbf{MMLU} & \textbf{Overall} &\textbf{Avg.} &\textbf{Std.} \\
    \Xhline{1px}
    $\text{Zephyr}^\dag$  & -  &UltraChat   &58.10 & 40.30 & 76.90 & 34.64 & 80.95 & 58.92 & 58.17  & - & -   \\
    $\text{Llama3-8B}^\dag$ & -  & -   &62.03 & 51.64 & 75.30 & 75.44 & 78.78 & 65.75 & 68.16 & - & -   \\
    $\text{Yi-6B}^\dag$   & -  & -   &57.51 & 50.01 & 71.98 & 40.63 & 78.48 & 63.17 & 60.30 & - & -   \\
    \Xhline{2\arrayrulewidth}
    Mistal-7B & SFT     & \multirow{4}{*}{UltraChat}   &63.31 & 49.13 & 78.77 & 42.53 & 83.79 & 62.04 & 63.26  &  63.34 & 0.09 \\
    Mistal-7B & PoFT    &    &63.40  & 49.46 & 78.77 & 44.88 & 83.83 & 62.10  & 63.74\textsubscript{$\uparrow$0.48} & 63.71\textsubscript{$\uparrow$0.38} & 0.04\\  
    \cdashline{1-2}\cdashline{4-12}
    Llama3-8B & SFT     &    &60.84 & 54.97 & 78.30 & 53.22 & 81.91 & 65.03 & 65.71 & 65.65   & 0.06 \\
    Llama3-8B & PoFT    &    &60.92 & 55.09 & 78.14 & 54.13 & 82.03 & 65.10 & 65.90\textsubscript{$\uparrow$0.19} & 65.88\textsubscript{$\uparrow$0.23} & 0.10 \\
    \hline
    Mistal-7B & SFT   & \multirow{4}{*}{\shortstack{Open-\\Hermes}}  &62.54 & 51.45 & 78.14 & 37.98 & 82.18 & 59.34 & 61.94 & 60.86 & 1.62 \\
    Mistal-7B & PoFT  &   &63.57 & 52.81 & 77.51 & 43.82 & 82.88 & 60.55 & 63.52\textsubscript{$\uparrow$1.58} &63.01\textsubscript{$\uparrow$2.15} & 0.50  \\ 
    \cdashline{1-2}\cdashline{4-12}
    Llama3-8B & SFT     &    &59.22 & 56.84 & 73.28 & 47.31 & 80.34 & 61.38 & 63.06 & 64.25  &0.81  \\
    Llama3-8B & PoFT    &    &61.43 & 58.38 & 75.77 & 50.72 & 81.63 & 63.42 & 65.23\textsubscript{$\uparrow$2.17} & 65.36\textsubscript{$\uparrow$1.11} &0.15  \\  
    \hline
    Mistal-7B & SFT   &\multirow{4}{*}{ShareGPT}   & 61.43 & 52.69 & 78.85 & 42.99 & 83.9  & 62.18 & 63.67 &63.66 &0.10 \\
    Mistal-7B & PoFT  &     & 61.86 & 52.74 & 78.45 & 45.19 & 84.02 & 62.21 & 64.08\textsubscript{$\uparrow$0.41}& 64.00\textsubscript{$\uparrow$0.34} &0.10 \\   
    \cdashline{1-2}\cdashline{4-12}
    Llama3-8B & SFT     &    &58.28 & 54.93 & 77.82 & 54.59 & 81.71 & 65.33 & 65.44 & 65.34  &0.11 \\
    Llama3-8B & PoFT    &    &57.76 & 55.07 & 78.30 & 55.95 & 81.75 & 65.15 & 65.66\textsubscript{$\uparrow$0.22} & 65.45\textsubscript{$\uparrow$0.11} &0.19 \\  
    \hline

  \end{tabular}
  \caption{Overall performance on LLM Open Leaderboard of Mistral-7B and Llama-3-8B training on UltraChat200k, OpenHermes, and ShareGPT. The last three columns present the results of the last epoch and the average scores and standard deviation across all epochs, respectively. We also present the results of the aligned LLMs, where $\text{Zephyr}^\dag$, $\text{Llama3-8B}^\dag$, and $\text{Yi-6B}^\dag$ stand for Zephyr-7B-sft-full, Llama-3-8B-Instruct, and Yi-6B-Chat respectively.}
  \label{overall_results}
\end{table*}

In this section, we present the main results of our experiments, highlighting the improvements achieved by PoFT across various datasets. Additionally, our ablation studies offer insights into the following aspects: (1) the effectiveness of PoFT on quality-limited data, (2) the comparison between PoFT and data filtering methods, and (3) the comparison between PoFT and data distillation from aligned LLMs.

\subsection{Settings}
\begin{itemize}
    \item \textbf{Training Data} To align with Zephyr-7B-sft-full \cite{zephyr}, we opt for UltraChat200k \cite{ultrachat} as the primary training dataset for PoFT. Besides, we also employ the ShareGPT-Chinese-English-90k dataset \cite{ShareGPT} and OpenHermes dataset \cite{OpenHermes}, which encompasses 240k data pairs. As ShareGPT comprises parallel bilingual data, we exclusively utilize the English corpus for training purposes.  Moreover, to examine PoFT's compatibility with DPO, we introduce the UltraFeedback \cite{ultrafeedback} dataset for two-step training.
    \item \textbf{Benchmarks} We evaluate models on the popular benchmark Huggingface Open LLM Leaderboard \cite{leaderborad}, MT-Bench \cite{mtbench} and AlpacaEval2.0 \cite{alpaca_eval}. Open LLM Leaderboard covers a variety of tasks, enabling the assessment of specific capabilities of LLMs. Both MT Bench and AlpacaEval 2.0 applied GPT-4 as the judge model to assess the model performance. The details of evaluation settings and metrics are listed in the Appendix \ref{sec:appendix2}. 
    \item \textbf{Model} We choose Mistral-7B-v0.1 \cite{mistral} and Llama-3-8B \cite{llama3} as backbones. For aligned LLMs, we adopt zephyr-7b-sft-full \cite{zephyr}, Llama-3-8B-Instruct \cite{llama3}, and Yi-6B-Chat \cite{yi}. Notably, Zephyr-7B-sft-full, derived from Mistral-7B-v0.1, trained on UltraChat200k. 
    
\end{itemize}

\subsection{Main Experiment}

We adopt the base models trained on the cross-entropy (CE) objective as our baseline (i.e., SFT model) and investigate the effectiveness of PoFT under the same training settings.
The experiments are conducted mainly on UltraChat200k, OpenHermes, and ShareGPT datasets.

Table \ref{overall_results} contains the experimental results of comparison between the models with different training objectives on the LLM Open Leaderboard. 
To ensure a fair evaluation, we report the results of the last epoch and the average scores across all training epochs after excluding the first epoch, which is typically considered unstable. 
Notably, as the same base model and datasets are used by Zephyr-7B-sft-full, by adjusting hyper-parameters, it could achieve better performance (see the fourth row of Table \ref{overall_results}).

Both scores on the Open LLM leaderboard show a consistent trend in which PoFT systematically outperforms the CE objectives across various training datasets on different base models. And the gap is more pronounced concerning OpenHermes, by 1.58 and 2.17 on Mistal-7B and Llama-3-8B, respectively.
Moreover, PoFT models have a comparable lower standard deviation than SFT models, indicating greater stability of PoFT across different training epochs.  
In terms of different evaluation datasets, PoFT models distinctly outperform SFT models on the GSM8k dataset.

\begin{table}[t]
\centering
\small
\begin{tabular}{l|l|lc|lc}
\Xhline{1px}
\multirow{2}{*}{\textbf{FT}} & \multirow{2}{*}{\textbf{Datasets}}    & \multicolumn{2}{c}{\textbf{MT-Bench}} & \multicolumn{2}{c}{\textbf{AlpacaEval(\%)}} \\

                       &                   & Last           & Avg.         & Last          & Avg          \\
\Xhline{1px}
Zephyr         & \multirow{3}{*}{UltraChat}                & 6.30          & -            & 3.91           & -             \\
SFT                    &   \multirow{2}{*}{}       & 6.35       & 6.05       & 3.98      & 3.93            \\
PoFT                &         & 6.52\textsubscript{$\uparrow$0.17}   & 6.12                &4.1\textsubscript{$\uparrow$0.12}  & 4.35                \\
\hline
SFT                    & \multirow{2}{*}{\shortstack{Open-\\Hermes}}   & 5.09  & 5.16      &4.86      & 3.99                 \\
PoFT                &     &5.93\textsubscript{$\uparrow$0.84}     &    5.89 &5.96\textsubscript{$\uparrow$1.1}   &   4.57      \\
\hline
SFT                    & \multirow{2}{*}{ShareGPT}  &6.63       & 6.44       &2.44   & 2.09      \\
PoFT                &    &6.83\textsubscript{$\uparrow$0.20}  & 6.60  &2.61\textsubscript{$\uparrow$0.17}  & 2.55         \\     
\hline 
\end{tabular}
\caption{Overall performance on MT-Bench and AlpacaEval 2.0 of Mistral-7B training on three datasets. The last two columns of each benchmark present the results of the last epoch and the average scores across all epochs, respectively. Specifically, the score for AlpacaEval 2.0 is the win rate(\%).} 
\label{overall_results2}
\end{table}

The results for MT-Bench and AlpacaEval 2.0 echo the findings from the LLM Open Leaderboard, with remarkable improvements being made in OpenHermes, shown in Table \ref{overall_results2}.
Nonetheless, the overall discrepancy between SFT and PoFT models is fairly minor. We attribute this to the evaluation perspectives of these benchmarks, such as helpfulness in human preference.


\begin{figure}[t]
    \centering
    \includegraphics[width=0.45\textwidth]{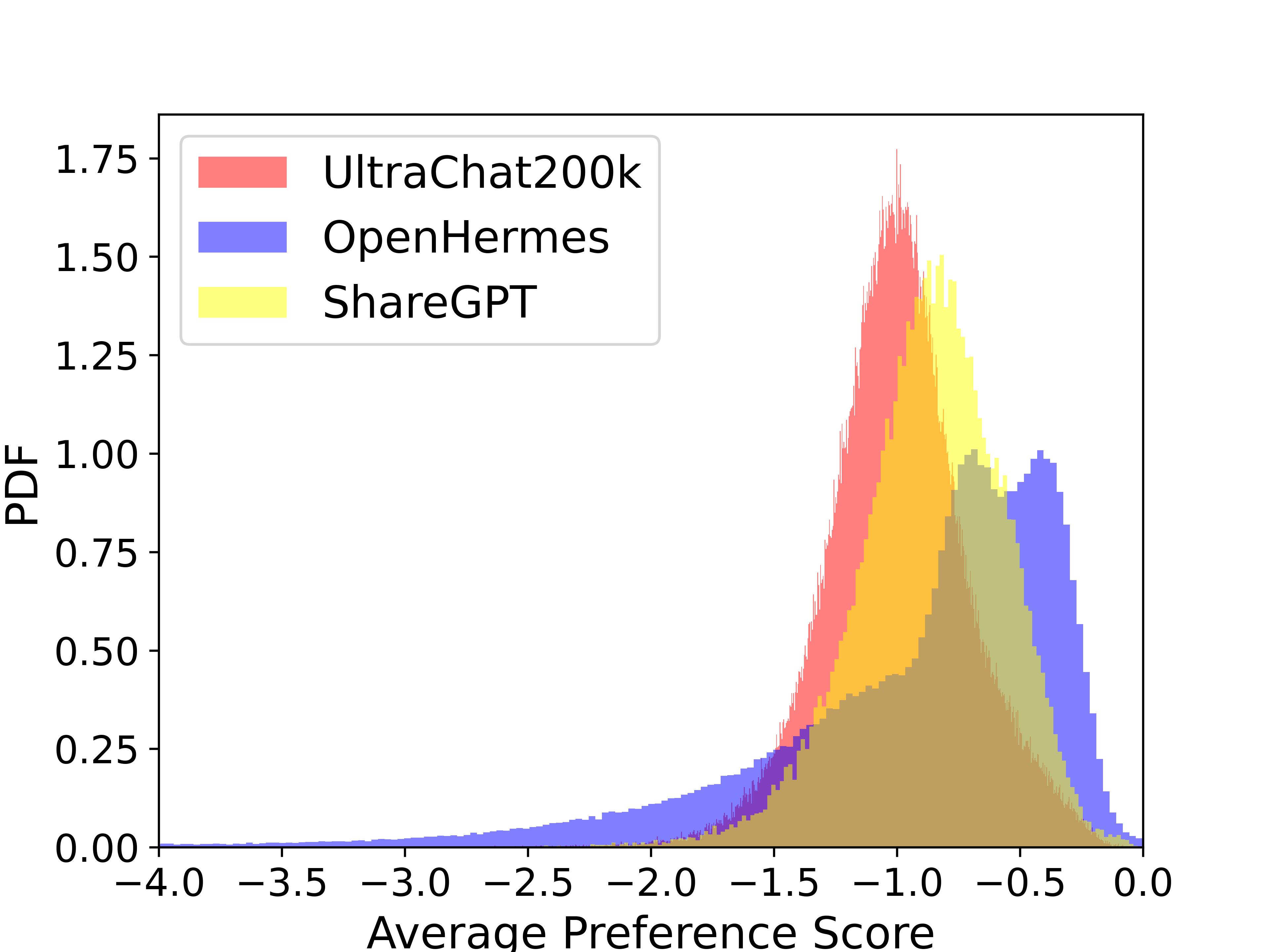}
    \caption{Preference scores generated by aligned LLMs across different training datasets. Note that PDF stands for the probability density function.}
    \label{reward_distribution}
\end{figure}

Since model performances on different training data are varied, we analyze the average preference scores distribution of aligned LLMs on three training datasets, depicted in Figure \ref{reward_distribution}. It is observed that the distributions of UltraChat200k and ShareGPT are more concentrated, while the distribution of OpenHermes is wider and flatter in shape. This implies the discrepancy of gradients on OpenHermes is more diverse during the training process, thereby amplifying the difference in training performance between CE and PoFT objectives. Hence, we can hypothesize that PoFT is inclined to a certain type of data distribution. In other words, under this data distribution,  our model can leverage its strengths more effectively than the SFT model. A comprehensive discussion regarding this observation is covered in section \ref{data_distribution}.


\begin{table*}[ht]
\centering
\scalebox{0.9}{
\begin{tabular}{l|l|lc|lc|lc}
\Xhline{1px}
\multirow{2}{*}{\textbf{Model}} & \multirow{2}{*}{\textbf{Datasets}}      & \multicolumn{2}{c}{\textbf{LLM Open Leaderboard}} & \multicolumn{2}{c}{\textbf{MT-Bench}} & \multicolumn{2}{c}{\textbf{AlpacaEval 2.0}} \\

                       &                                & Avg.                 & Std.                & Avg.           & Std.          & Avg.(\%)           & Std.           \\
\Xhline{1px}
Zephyr\small{+DPO}              &  \multirow{3}{*}{\makecell[c]{+UltraFeedback}}                          & 62.62               & -                  & 7.11          & -            & 19.01           & -             \\
SFT\small{+DPO}                    &     & 65.18              & 0.32               & 6.84          & 0.25         & 25.09           & 1.72          \\
PoFT\small{+DPO}                     &      & 65.88 \textsubscript{$\uparrow$0.70}                & 0.12               & 7.04\textsubscript{$\uparrow$0.20}           & 0.08         & 27.83\textsubscript{$\uparrow$2.74}            & 3.06          \\
\hline 
\end{tabular}}
\caption{Performance of two-step training models based on Mistral-7B. Specifically, the average score for AlpacaEval 2.0 is the average win rate(\%). For comparison, we also present the results of Zephyr-7b-beta, denoted as Zephyr+DPO.} 
\label{dpo_results}
\end{table*}

In addition to comparing our approach with SFT methods, we also investigate the compatibility between PoFT and DPO. The results in Table \ref{dpo_results} demonstrate that integrating PoFT and DPO can yield superior performance across all benchmark tasks. It is worth noticing that this combined approach brings a significant improvement on the AlpacaEval benchmark, with the win rate surging to 27.83\%, underscoring the effectiveness of PoFT-DPO synergy.

\subsection{Ablation Study}
\begin{figure*}[ht]
    \centering
    \scalebox{1}{
    \subfloat[][]{\includegraphics[width=0.32\textwidth]{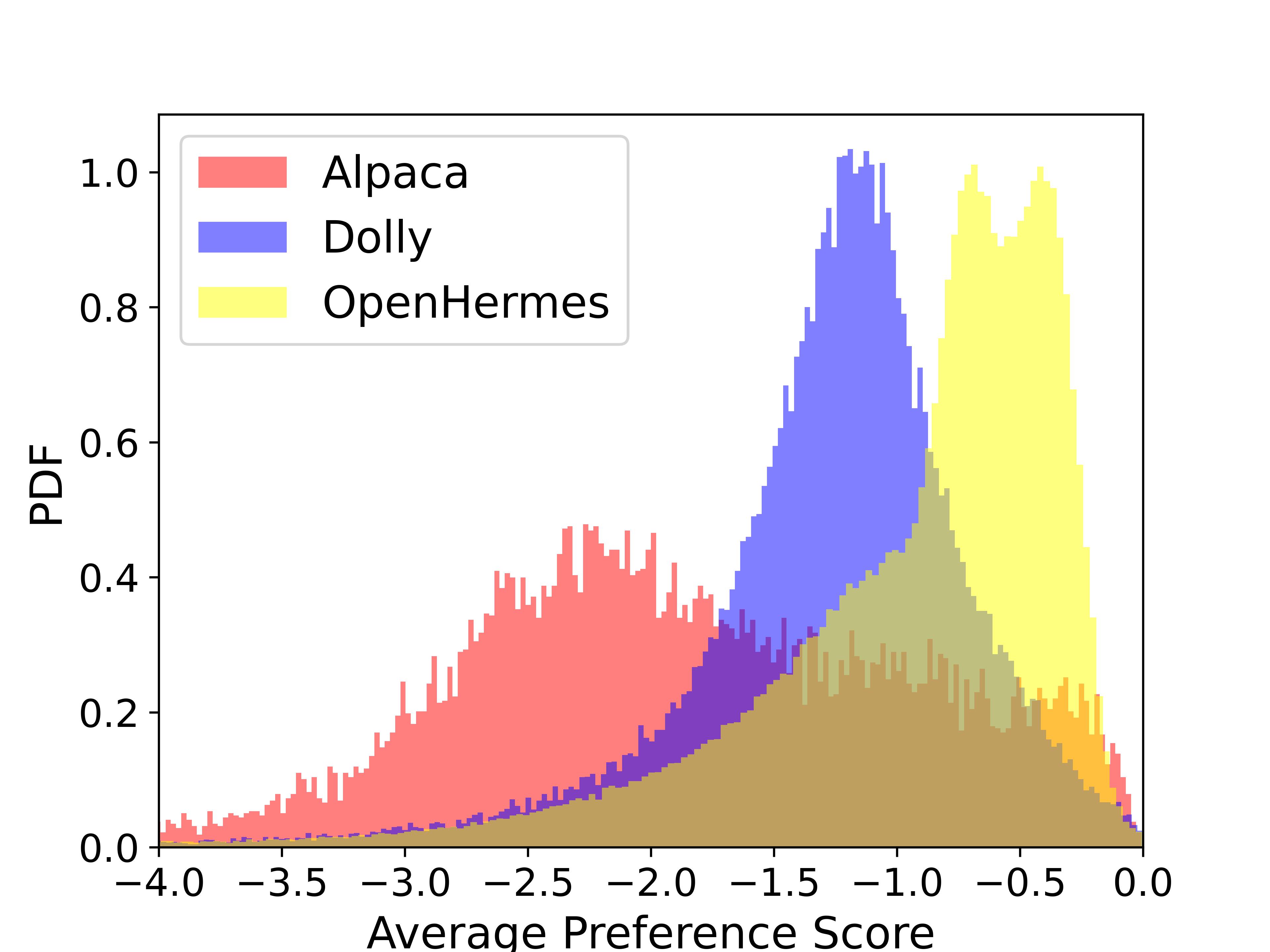}\label{alpaca_dolly_distribution}}
    \hfill
    \subfloat[][]{\includegraphics[width=0.32\textwidth]{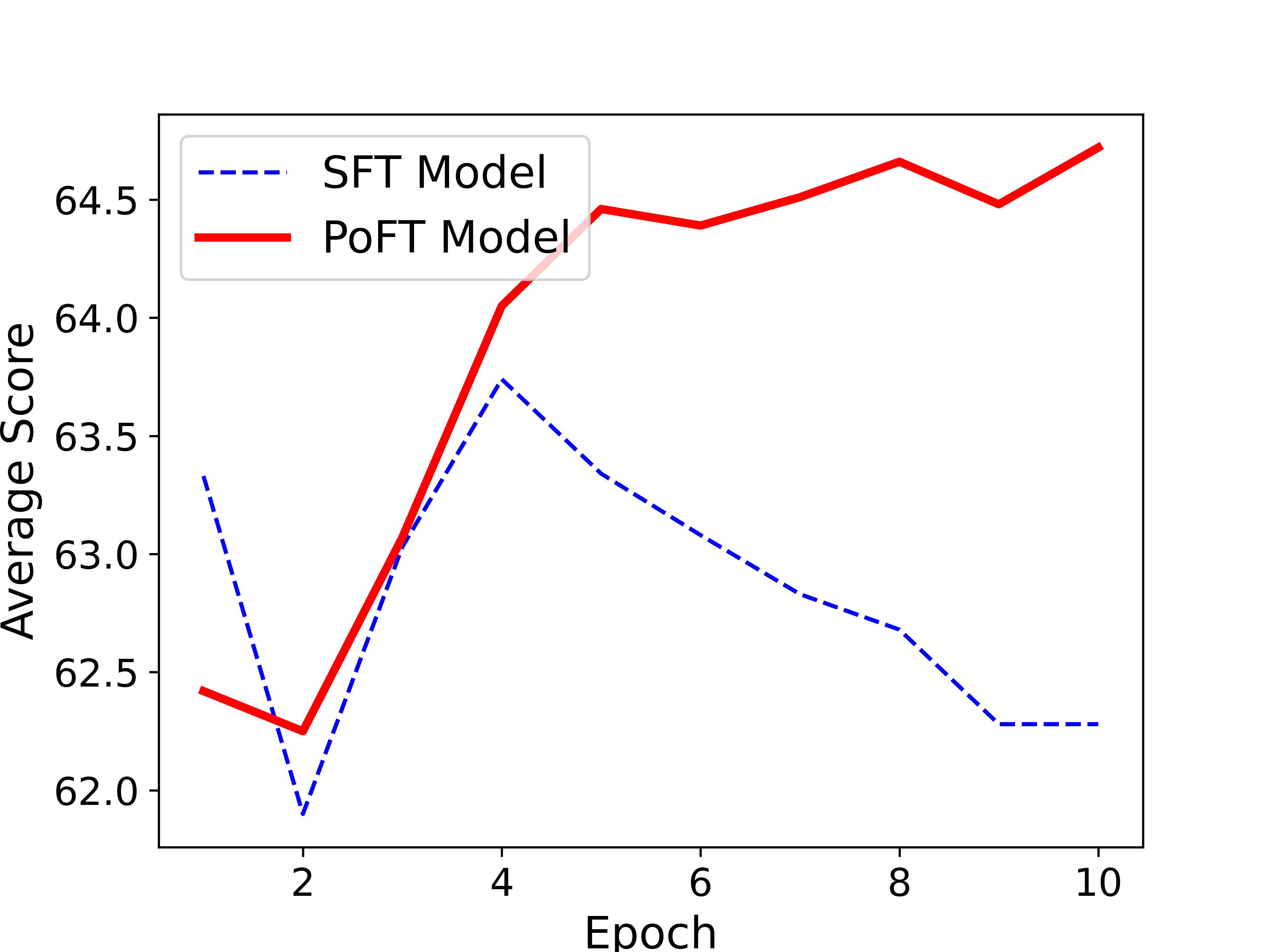}\label{time_step_alpaca}}
    \hfill
    \subfloat[][]{\includegraphics[width=0.32\textwidth]{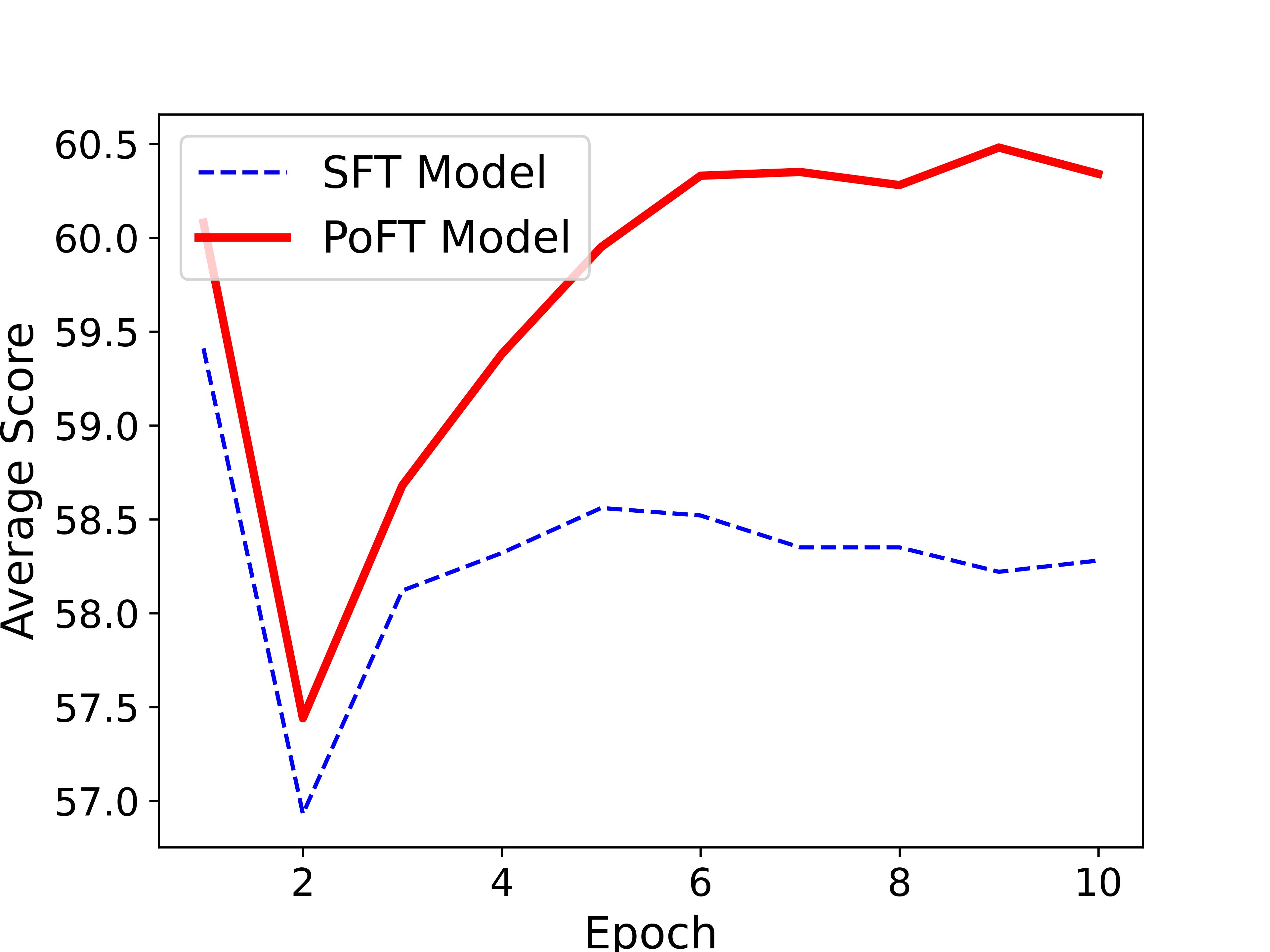}\label{time_step_dolly}}}
    
    \vspace*{1pt} 
    
    \begin{minipage}{\textwidth}
      \centering
      \scalebox{1}{
      \subfloat[][]{\includegraphics[width=0.33\textwidth]{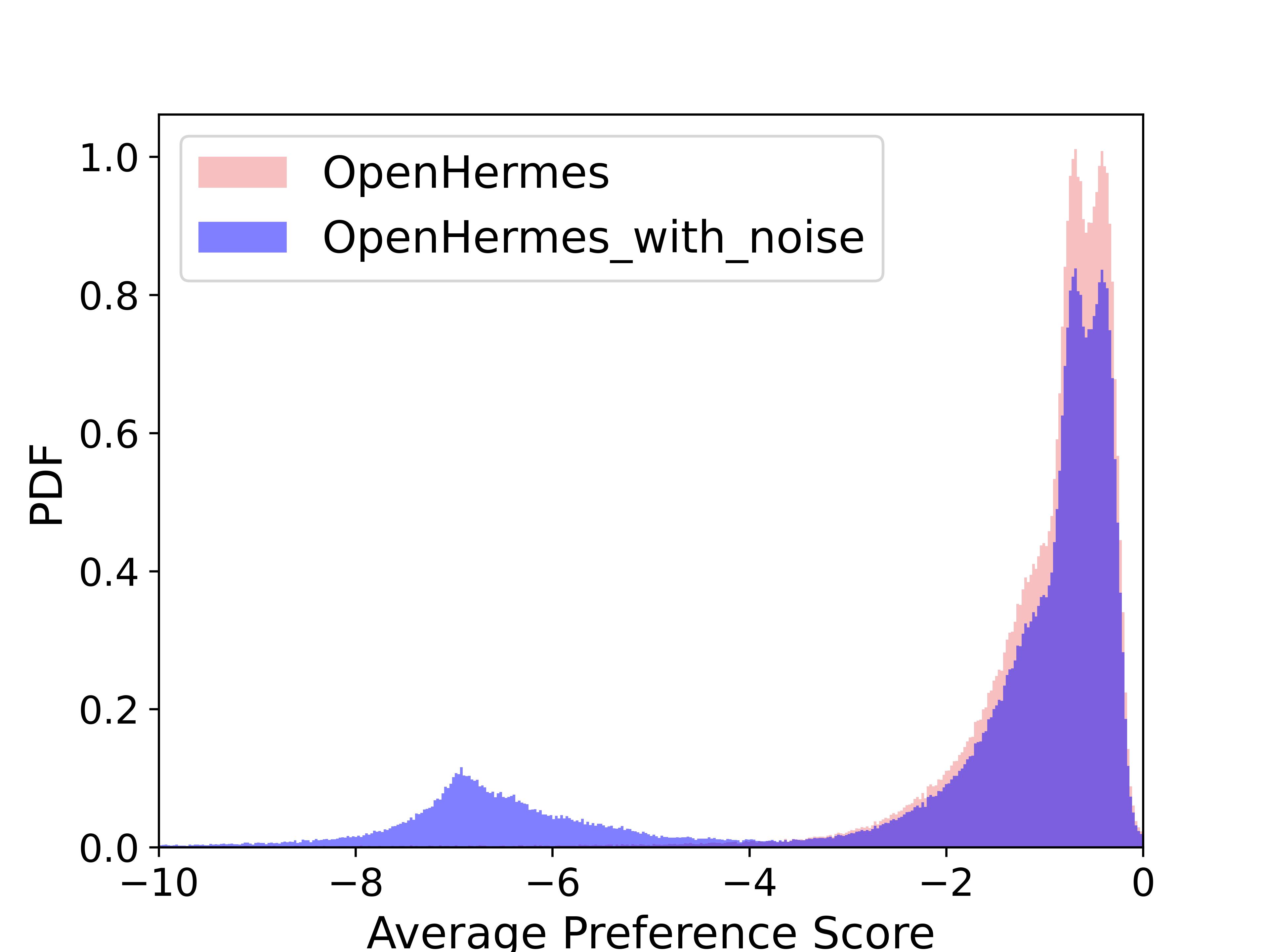}\label{noise_distribution}}
      \hspace{15pt} 
      \subfloat[][]{\includegraphics[width=0.33\textwidth]{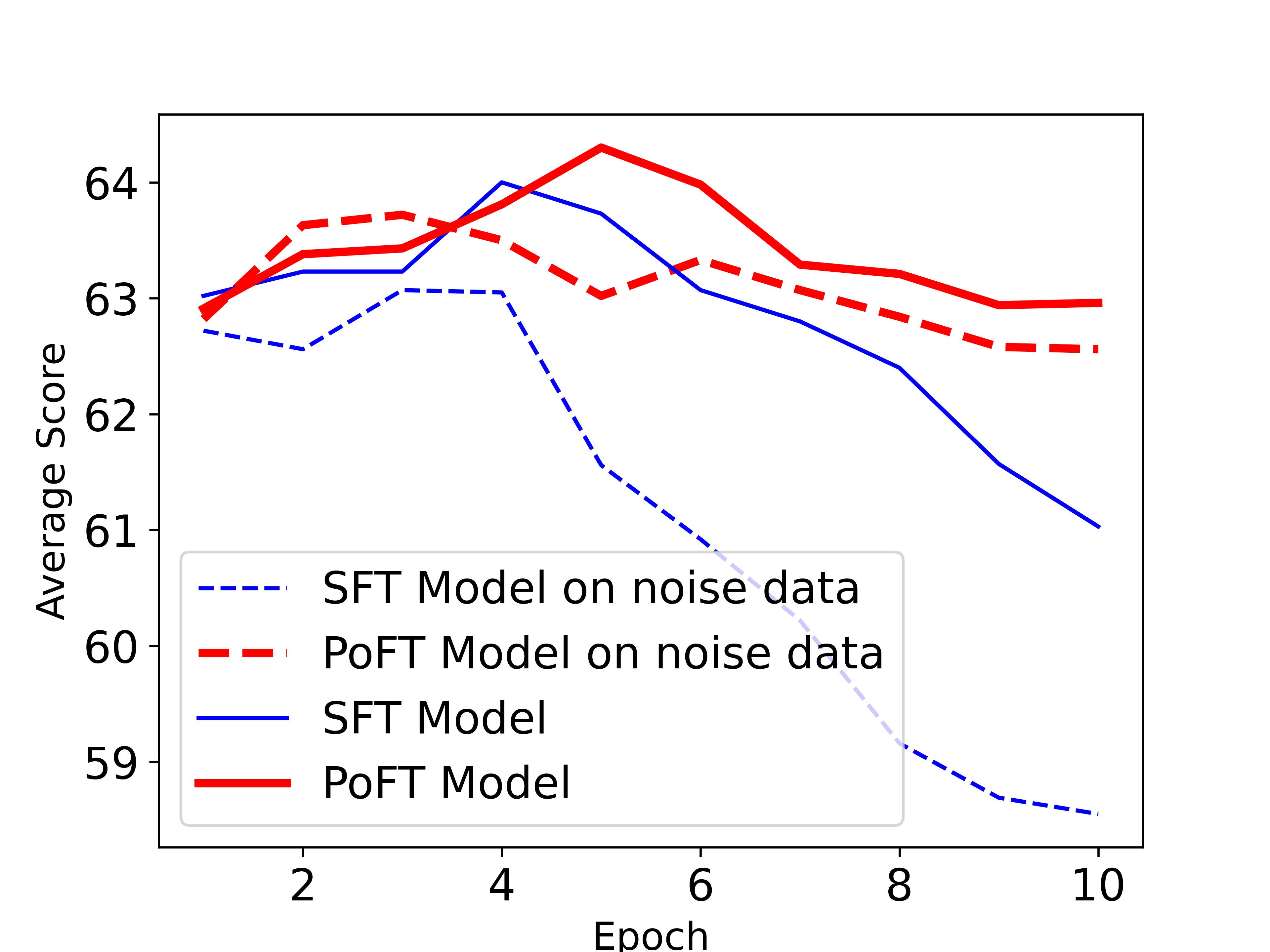}\label{time_step}}}
    \end{minipage}
    \caption{Analysis and model performances on quality-limited data. (a) Preference score distributions of data-limited datasets - Alpaca and Dolly, compared to OpenHermes. Note that PDF stands for the probability density function. (c) Performances of PoFT and SFT models training with Alpaca. (d) Performances of PoFT and SFT models training with Dolly. (d) Preference score distributions of hand-crafted noise data on OpenHermes. The increase in the long-trail part indicates the distribution of the noise data. (e) Performances of PoFT and SFT models training with hand-crafted noise data.}
\end{figure*}

\subsubsection{Effectiveness on Quality-limited Data}\label{data_distribution}



The Cross-Entropy (CE) objective is vulnerable to poor data quality as it does not differentiate between high and low-quality data.
In contrast, by integrating aligned LLMs, PoFT can diversify the impacts of data during the optimization process.


However, there is a disparity in the improvements of PoFT for different datasets. By observing the preference score distribution in Figure \ref{reward_distribution}, we assume that this disparity could be attributed to the distribution of training data. 
Intuitively, when the distribution is highly concentrated, the gap between SFT and PoFT diminishes as the weights for different samples are less diverged. This leads us to our assumption that, upon training on a dataset with a more diverse preference score distribution, a more significant enhancement in PoFT could be observable over the SFT model. 

To verify our assumption, we conduct experiments on the datasets Alpaca \cite{alpaca} and Dolly \cite{dolly}, which are regarded as quality-limited datasets \citep{backtranslation,instag}. Note that we intentionally increase the number of training epochs to ten for a more nuanced observation of the effects over an extended period.
Figure \ref{time_step_alpaca} and Figure \ref{time_step_dolly} depict the performance of Mistral-7B models training with these two datasets respectively. During the initial epochs, there is a significant drop in both models. We attribute this to the significant percentage of noise data within the datasets. Nevertheless, PoFT is more robust, proven by the consistent improvement over subsequent epochs. Meanwhile, SFT models are under-performed, indicated by a decreasing trend.

To present the noise data more intuitively, we directly construct hand-crafted noise data to increase the data in the long-tail part of the preference score distribution.  Utilizing OpenHermes as our source, we create a pair of inputs with a randomly matched output and simulate data corruption through the processes of character insertion, deletion, and modification, yielding 50k noise data. The newly created noise data is blended with the original data for training. Figure \ref{noise_distribution} presents the distribution of new training data. 

Figure \ref{time_step} elaborates the performance of Mistral-7B trained on the noise data. For comparison, we also display the performance of models trained on the original data under the same training settings.  
Overall, the PoFT models consistently surpass the performance of the SFT models regardless of the data settings. It is worth noting that the gap between the PoFT and SFT models is widened when trained with the noise data. As the training epoch increases, there is a remarkable drop in SFT models, particularly for the one with noise data. This indicates that SFT training is more likely to result in over-fitting, which is exacerbated by the noise in data. In contrast, PoFT shows impressive stability.

The studies above underscore the resilience of PoFT in dealing with various data qualities, which can be attributed to the preference scores from aligned LLMs. These scores help mitigate the negative effects of noisy data, emphasizing the higher-quality data during optimization, leading to a significant improvement.


\subsubsection{PoFT v.s. Data Filtering}\label{data_filtering}

When associating with the reward function, the coefficient in Eq.\ref{coefficient} can be interpreted as:

\begin{equation}
    \frac{\exp(\frac{1}{M}\sum_{j=1}^{M}r_j(x,y))}{\exp(\frac{1}{M}\sum_{j=1}^{M}r_j(x,y))+\exp(r_\theta(x,y))},
\end{equation}
where $r_\theta(x,y)$ and $r_j(x,y)$ refer to the rewards (i.e., preference scores) of the target model and $j$-th aligned LLM, respectively, and $M$ is the number of aligned LLMs.
Intuitively, the preference scores assigned by aligned LLMs could directly guide the optimization process -- higher scores increase gradient update weight. 
As $r_j(x,y)$ dynamically affects the importance of the samples in training, the PoFT objective can be regarded as a soft filtering approach.


To evaluate this implicit data-filtering mechanism, we apply preference scores to filter data directly. In our experiment, we first sort the data by the scores in descending order. Subsequently, we set thresholds to select varying percentages of data and train PoFT and SFT objectives accordingly.

\begin{figure}[ht]
    \centering
    \includegraphics[width=0.45\textwidth]{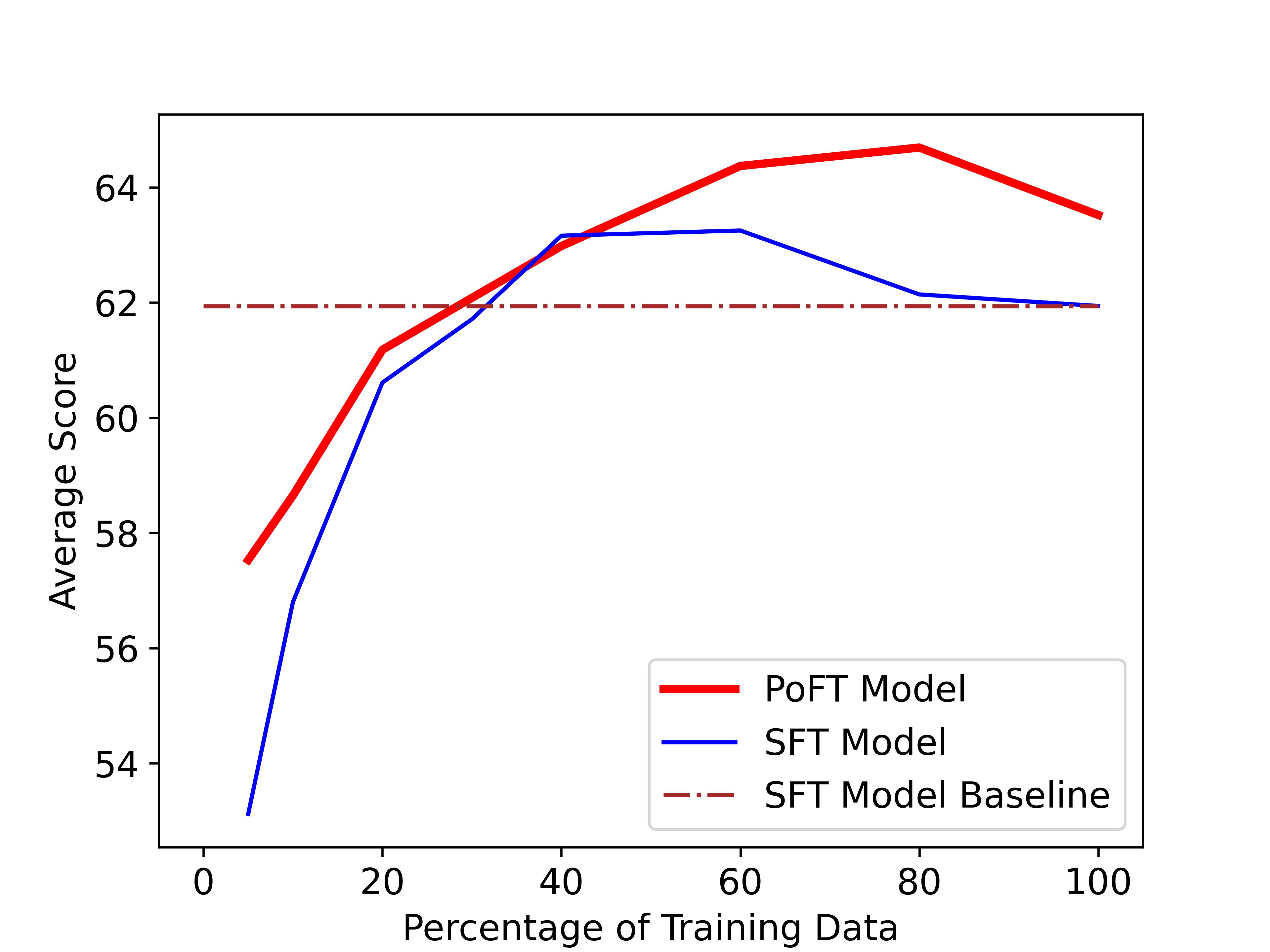}
    \caption{Performance of Mistral-7B trained with different percentages of data on Open LLM Leaderboard. }
    \label{percent_data}
\end{figure}

Figure \ref{percent_data} demonstrates the model performances on the filtered data. In the initial stages, as the number of data increases, there is a positive trend on the SFT model, peaking at 40 percent. This performance even surpasses that of the model trained on the entire dataset. However, despite the continued increase in data volume, the performance begins to decline as more data of inferior quality are included. Interestingly, when trained on filtered data, the PoFT model can further enhance performance.


This steers us toward the hypothesis that combining PoFT and other filtering methods could further enhance overall performance. We assume that PoFT, when employed in conjunction with other filtering strategies, can deliver a multifaceted evaluation of data quality, resulting in a more comprehensive filtering process.
Therefore, we conduct experiments on the widely recognized SFT-filtering techniques -- IFD \cite{IFD}, Instag \cite{instag}, and Deita \cite{makesgooddataalignment}. In detail, these filtering methods are applied to the OpenHermes dataset to filter out 20\% of data. Subsequently, the Mistral-7B models are trained with CE and PoFT objectives on these data. 

The overall performance is illustrated in Table \ref{filtering_compare}. It is indisputable that these filtering methods significantly enhance the performance of SFT, even surpassing PoFT models with full data training. Nonetheless, the utility of these methods is not in contention with our approach. In fact, they can be seamlessly integrated with PoFT, yielding performance superior to applying either method in isolation.

\begin{table}[ht]
\centering
\scalebox{0.9}{
\begin{tabular}{l|l|c}
\hline
Filtering method            & FT   & Overall       \\
\hline
\multirow{2}{*}{N/A}        & SFT  & 61.94         \\
                            & PoFT & 63.52    \\
\hdashline
\multirow{2}{*}{Preference score} & SFT  & 62.14          \\
                            & PoFT & \textbf{64.69}   \\
\hline
\multirow{2}{*}{IFD \cite{IFD}}        & SFT  & 64.71          \\
                            & PoFT & \textbf{64.95} \\
\hline
\multirow{2}{*}{Instag \cite{instag}}     & SFT  & 64.04          \\
                            & PoFT & \textbf{64.27} \\
\hline
\multirow{2}{*}{Deita \cite{makesgooddataalignment}}     & SFT  & 64.31          \\
                            & PoFT & \textbf{64.49} \\
\hline
\end{tabular}}
\caption{Performance of Mistral-7B models trained with filtered data on Open LLM leaderboard. We present the overall results of the last epoch.}
\label{filtering_compare}
\end{table}

In summary, the experiments confirm: (1) our reward function is effective since using preference scores for filtering allows the model to achieve superior performance on less amount of data; (2) PoFT is compatible with other data filtering methods, further enhancing the overall performance.



\subsubsection{PoFT v.s. Data Distillation From Aligned LLMs}

The commonality between PoFT and data distillation is that they both leverage additional LLMs to provide information for model training. However, PoFT incorporates aligned LLMs to guide the gradient optimization process via preference modeling, while data distillation aims at transferring knowledge from the teacher models, rather than solving problems regarding the data quality.

To compare PoFT and data distillation methods, we employ aligned LLMs as teachers to create the responses of OpenHermes, resulting in a new training set. The experiments are conducted on these synthesized data with the CE objective. Intuitively, regenerating the responses is a more explicit way to amplify the effectiveness of aligned LLMs.


\begin{table}[ht]
\centering
\scalebox{0.9}{
\begin{tabular}{l|l|cc}
\hline
FT                & Regen-Model        & Overall     \\
\hline
\multirow{4}{*}{SFT} & N/A               &61.94  \\
                    & Llama-3-8B-instruct &63.16  \\
                    & Zephyr-7B-sft-full &62.30   \\
                    & Yi-6B-Chat         &60.18   \\
\hdashline
PoFT             & N/A                & \textbf{63.52}  \\
\hline
\end{tabular}}
\caption{Performance of Mistral-7B models training with regenerated data on Open LLM Leaderboard. We present the results of the last epoch.}
\label{regeneration_comparison}
\end{table}

The results on the LLM Open Leaderboard are presented in Table \ref{regeneration_comparison}. Surprisingly, directly replacing the original responses with synthesized data leads to performance degradation. The models trained on the regenerated data underperform PoFT models, performing even worse than the SFT model trained on the original data in some cases. The decrease is more remarkable in the teacher model Yi-6B-Chat.  

To sum up, although directly applying aligned LLMs for data regeneration is a more straightforward way for incorporation, it could introduce variability and uncertainty, degrading the model performance. Hence, PoFT offers a more appropriate way of incorporation, efficiently taking advantage of those aligned LLMs through preference modeling.


\section{Conclusion}
In this paper, we present PoFT, a novel and effective preference-oriented SFT method by applying the Bradley-Terry objective for modeling preferences between different models. 
Specifically, given the same SFT data, we intentionally define a preference: favoring the target model over aligned LLMs.
This preference encourages the target model to generate higher preference scores when compared to the aligned LLMs.
In essence, the aligned LLMs provide assessments of the data quality in the optimization process, varying the effects of SFT data.
We conduct extensive experiments on diverse training datasets and different base models to verify the efficacy of PoFT compared to the baselines (the CE objective). 
Furthermore, we prove its stability towards noise data and validate the effectiveness of the designed objectives by conducting ablation studies on the reward functions and aligned LLMs. 
Furthermore, PoFT can be combined with other SFT Filtering methods to attain enhanced performance outcomes.
Notably, integrating PoFT with DPO has the potential to yield even superior performance.


\appendix

\section{Math Deduction}
\label{sec:appendix1}

Eq.\ref{ce_grad_details} and Eq.\ref{PoFT_grad_details} show gradient derivation process of cross-entropy and PoFT loss respectively:

\begin{equation}
    \begin{split}
    \nabla_\theta L_{ce} &= \nabla \left[ -\frac{1}{T_0(y)} \log p_\theta\left(y \mid x \right)\right]  \\
                    &= - \frac{1}{T_0(y)} \frac{1}{p_\theta\left(y \mid x \right)} \nabla  p_\theta\left(y \mid x \right)  \\
    \end{split}
    \label{ce_grad_details}
\end{equation}

\begin{equation}
    \begin{split}
    \nabla_\theta L_{\text{PoFT}} &= \nabla \left[ -\log \sigma  \left(\frac{1}{M}\sum_{j=1}^M  \left( \frac{1}{T_0(y)}\log p_\theta(y \mid x) \right.\right.\right.\\ &\left.\left.\left. -\frac{1}{T_j(y)}\log p_j(y \mid x)  \right)\right) \right] \\
                              &= \nabla \left[ -\log \frac{p_\theta(y \mid x)^\frac{1}{{T}_0(y)}}{p_\theta(y \mid x)^\frac{1}{{T}_0(y)}+ \tau} \right]  \\
                              &= - \frac{1}{{T}_0(y)} \frac{1}{p_\theta\left(y \mid x \right)} \tau \nabla  p_\theta\left(y \mid x \right), \\
    \end{split}
    \label{PoFT_grad_details}
\end{equation}
where
\begin{small}
\begin{align*}
    \tau =  \frac{\left( \prod \limits_{j=1}^M p_j(y \mid x)^\frac{1}{{T}_j(y)} \right)^\frac{1}{M}}{\left( \prod \limits_{j=1}^M p_j(y \mid x)^\frac{1}{{T}_j(y)} \right)^\frac{1}{M}+p_\theta(y \mid x)^{\frac{1}{{T}_0(y)}}}.\\
\end{align*}
\end{small}

\section{Setup Details}
\subsection{Training Settings}
We utilize eight NVIDIA Tesla A100 GPUs to train Mistral-7B and Llama-3-8B models. All training is based on alignment-handbook \cite{alignment_handbook2023}. 
For parallel training, we utilize FlashAttention2 \cite{flashattention} and DeepSpeed Zero-Stage 3 \cite{zero}. 
The sequence length and warm-up rate are set to 2048, 4096, and 10\%, respectively. 
To explore better performance, learning rate is varied from 5e-6 to 5e-5. The model is trained for three to four epochs, with a cosine warmup scheduler.
When training with multi-turn corpus, we apply the corresponding chat templates for the base model, i.e., Zephyr's template for Mistral-7B and Llama-3-8B-Instruct's template for Llama-3-8B.
As for the DPO training, we set a learning rate of 5e-7 over two epochs with a linear warmup scheduler, aligning with Zephyr-7B-beta \cite{zephyr},

\subsection{Evaluation Metric}
\label{sec:appendix2}

\begin{table}[ht]
\small
\begin{tabular}{llll}
\hline
\textbf{datasets}  & \textbf{Arc} & \textbf{TruthfulQA} & \textbf{Winogrande} \\
\hline
Few-shot          &   25        &  0   &  5     \\
Metric            &  acc\_norm   & acc  & acc \\
\hline
\textbf{datasets} & \textbf{GSM8k} & \textbf{HellaSwag}  & \textbf{MMLU} \\
\hline
Few-shot          &  5   &  10    &   5   \\
Metric            & acc(flexible)    &  acc\_norm    & acc\_norm \\
\hline
\end{tabular}

\caption{Details of evaluation implementation on HuggingFace Open LLM Leaderboard. Specifically, we show the few-shot setting and evaluation metric for each dataset.}
\label{eval_matrix}
\end{table}

LLM Open Leaderboard comprises Arc \cite{arc}, TruthfulQA \cite{TruthfulQA}, Winograde \cite{WinoGrande}, GSM8k \cite{gsm8k}, HellaSwag \cite{HellaSwag}, and MMLU \cite{mmlu}. The detailed evaluation settings for each dataset are presented in Table \ref{eval_matrix}. The evaluation is conducted on lm-evaluation-harness Library \cite{eval-harness} with the version of 0.4.2.\footnote{https://github.com/EleutherAI/lm-evaluation-harness}
MT-Bench consists of 80 questions within 8 categories, requiring multi-turn conversation ability. AlpacaEval 2.0 involves 805 questions, over which we compute the win rate of the target model against the reference model, specifically gpt4\_turbo in our context.
In terms of leaderboards with extra judge models, we utilized gpt-4-0613 and gpt-4-1106-preview for MT-Bench and AlpacaEval 2.0 respectively. MT-Bench is conducted on FastChat Library\footnote{https://github.com/lm-sys/FastChat/tree/main}, while AlpacaEval is conducted on alpaca\_eval Library\footnote{https://github.com/tatsu-lab/alpaca\_eval}.

\section{Complementary Experiments}
\label{sec:appendix3}

\subsection{The Choice of Aligned LLMs}
The preference scores of aligned LLMs are decided by three aligned LLMs in our context. Similar to majority voting, the collaborative scores are more robust and stable. 
To examine this assumption, we conduct experiments on different choices of the aligned LLMs by changing the collaborative preference scores to individual ones, e.g. only using the preference scores generated by Llama-3-8B-Instruct.

\begin{table}[ht]
\small
\centering
\begin{tabular}{l|l|ccc}
\hline
FT                 & LLM(s) & UltraChat  & OpenHermes     & ShareGPT \\
\hline
SFT                   & -           & 63.34          & 61.94          & 63.66    \\
\hdashline
\multirow{4}{*}{PoFT} & ALL         & \textbf{63.71}          & 63.52          & 64.00    \\
                      & Llama3 & 63.55          & \textbf{64.29} &    64.07 \\
                      & Zephyr & 63.65          & 64.18          & 64.07    \\
                      & Yi     & 63.59          & 62.95          & \textbf{64.13}    \\
\hline
\end{tabular}
\caption{Performance of different choices of aligned LLMs on LLM Open Leaderboard. We present the performance of the last epoch. The best results under different training datasets are in bold.}
\label{choice_of_llms}
\end{table}

The results of PoFT models training with different aligned LLMs are shown in Table \ref{choice_of_llms}. Compared to SFT models, there are significant improvements in PoFT models regardless of aligned LLM choices. Moreover, in some cases, models using a single aligned LLM underperform the collaborative ones, indicating the instability of single decisions. Although the model with single aligned LLM has the best performance across different datasets, the best choices on different training datasets are varied, making it hard to select the best aligned LLM. Thus, in our experiments, we leverage the collaborative scores to obtain a more stable performance. More explorations of the aligned LLMs choice are left for future work.

\subsection{Strategies for Integrating Aligned LLMs}

In our approach, we generally prefer to combine the preference scores from various aligned Large Language Models (LLMs) by calculating their average, which serves to provide a collaborative estimate for a specific sample $\langle x,y \rangle$.
Yet, we also wish to explore varying ways to amalgamate these aligned LLMs to arrive at a more reliable conclusion.
Consequently, akin to our main experiments, we scrutinize various integration strategies such as \textit{Max} and \textit{Min} applied to the preference scores of Aligned LLMs, i.e. $\max\{r_{\text{LLM}_1}(x,y)\,\dots\,r_{\text{LLM}_j}(x,y)\}$ and $\min\{r_{\text{LLM}_1}(x,y)\,\dots\,r_{\text{LLM}_j}(x,y)\}$ respectively.

\begin{table}[ht]
\small
\centering
\begin{tabular}{l|l|ccc}
\hline
FT                 & Strategy & UltraChat      & OpenHermes     & ShareGPT       \\
\hline
SFT                   & -        & 63.34          & 61.94          & 63.67          \\
\hdashline
\multirow{3}{*}{PoFT} & Avg.     & \textbf{63.71}          & 63.52          & \textbf{64.08} \\
                      & Min.     & 63.61 & 64.13          & 64.05          \\
                      & Max.      & 63.64          & \textbf{64.30} & 63.91          \\
\hline
\end{tabular}
\caption{Performance of different integration strategies on LLM Open Leaderboard. We present the performance of the last epoch. The best results under different training datasets are in bold.}
\label{intergrating_strategies}
\end{table}

Table \ref{intergrating_strategies} illustrates the performance of different integration strategies. Notably, PoFT models with different strategies all surpass the SFT models.
Nonetheless, the performance gains brought about by these strategies appear to be inconsistent, varying across different training datasets. 
Despite the fluctuating results, the \textit{Avg.} strategy typically offers superior performance compared to its counterparts. 
Hence, we opt for the average strategy in the main sections of our analysis, balancing considerations of performance and generalizability.

\subsection{PoFT For Binary Preference Modeling}
We impose a specific preference in Eq.\ref{eq1} by requiring $\mathcal{P}\longrightarrow 1$, effectively prompting the target model to generate higher preference scores than the aligned LLMs. 
This preference uniformly applies to all SFT data without making distinctions based on data quality.

However, in section \ref{data_filtering}, the noise data are intentionally created and can be distinguished from pure SFT data.
By incorporating this label information, we can convert the objective into the subsequent form for binary preference modeling, hereafter referred to as bi-PoFT:
\begin{small}
\begin{equation}
\begin{split}
& L_{\text{bi-PoFT}}(\boldsymbol{\theta}) \\ 
& = -\mathbb{E}_{(x,y)\sim \mathcal{D}_{\text{SFT}^+}, r_\theta(x,y) \succ  r_{\text{LLMs}}(x,y)  \sim \mathcal{P}(\cdot)} \left[\log \mathcal{P}_\theta(\cdot)\right]\\
&-\mathbb{E}_{(x,y)\sim \mathcal{D}_{\text{SFT}^-}, r_{\text{LLMs}}(x,y) \succ r_\theta(x,y) \sim 1-\mathcal{P}(\cdot)} \left[\log \left(1-\mathcal{P}_\theta(\cdot)\right)\right]\\
& \approx -\mathbb{E}_{(x,y) \sim \mathcal{D}_{\text{SFT}^+}} \left[\log \frac{\exp \left(r_\theta(x,y)\right)}{\exp \left(r_\theta(x,y)\right)+ \exp\left(r_{\text{LLMs}}(x,y)\right)}\right]\\
& -\mathbb{E}_{(x,y) \sim \mathcal{D}_{\text{SFT}^-}} \left[\log \frac{\exp\left(r_{\text{LLMs}}(x,y)\right)}{\exp \left(r_\theta(x,y)\right)+ \exp\left(r_{\text{LLMs}}(x,y)\right)}\right]\\
& =  -\mathbb{E}_{(x,y) \sim \mathcal{D}_{\text{SFT}^+}} \left[\log \sigma  \left(\frac{1}{M}\sum_{j=1}^M  \left( r_\theta\left(x,y\right) -r_j\left(x,y\right)\right)\right)\right]\\
&-\mathbb{E}_{(x,y) \sim \mathcal{D}_{\text{SFT}^-}} \left[\log \sigma  \left(\frac{1}{M}\sum_{j=1}^M  \left( r_j\left(x,y\right) -r_\theta\left(x,y\right)\right)\right)\right],
\label{bi_poft}
\end{split}
\end{equation}
\end{small}
where $\mathcal{D}_{\text{SFT}^+}$ and $\mathcal{D}_{\text{SFT}^-}$ refer to the general SFT data and hand-crafted noise SFT data, respectively. 
Note that $\mathcal{P}\longrightarrow 1$ for general SFT data and $\mathcal{P}\longrightarrow 0$ for noise data.
In this objective, we define two different preferences for these two types of data: $r_\theta(x,y) \succ r_{\text{LLMs}}(x,y) $ and $r_{\text{LLMs}}(x,y) \succ r_\theta(x,y)$, where the aligned LLMs still serve as a borderline for the target model.
In contrast to the patterns on the general data, this objective prompts the target model to generate lower preference scores than that generated by aligned LLMs on the noise data, following the distribution $1-\mathcal{P}(\cdot)$.
This approach effectively emphasizes the difference in the optimization trajectories for general and noise-specific SFT data, thereby enhancing the overall optimization process.

\begin{figure}
    \centering
     \includegraphics[width=0.95\linewidth]{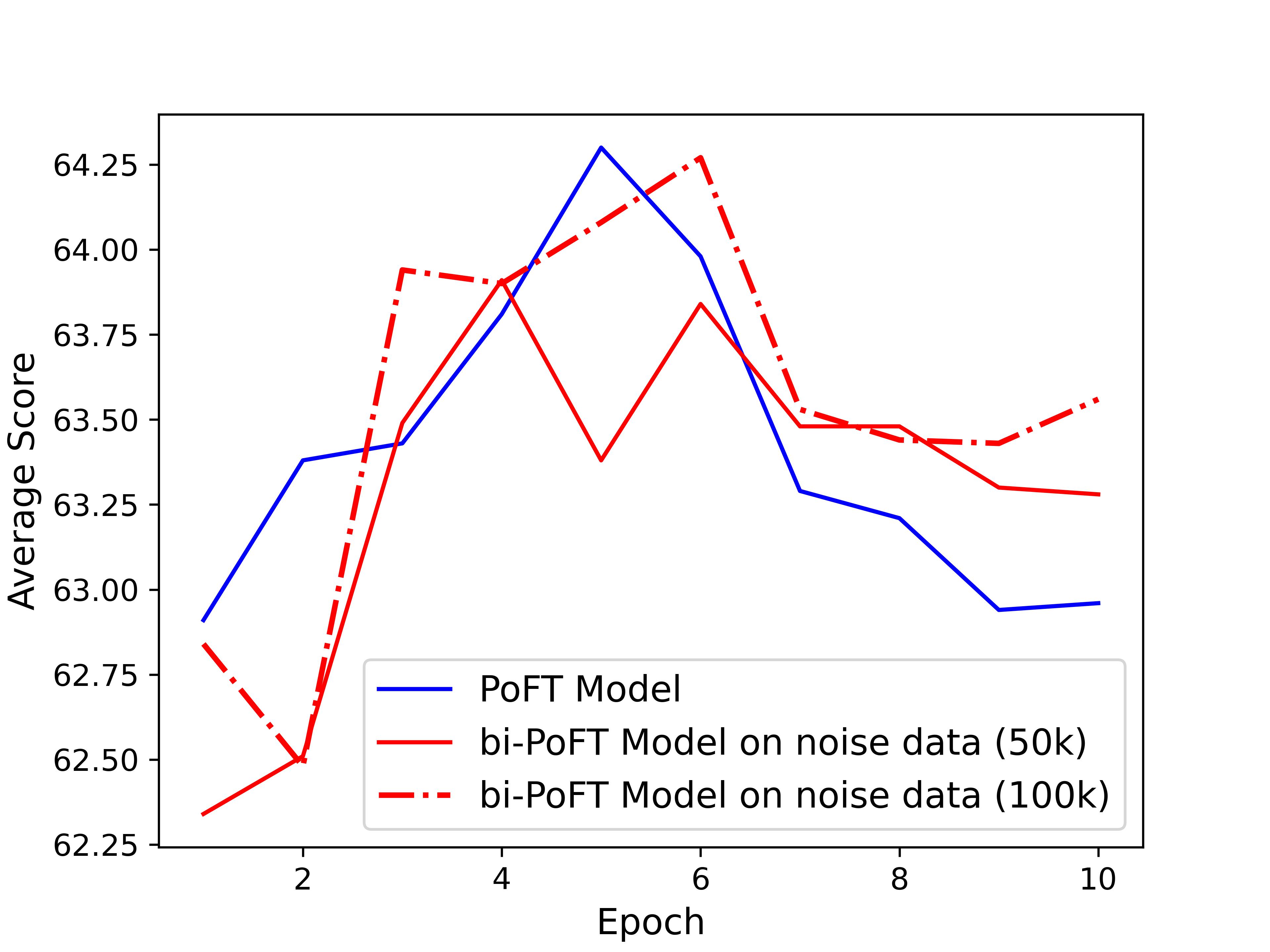}
    \caption{Performances of bi-PoFT models training with hand-crafted noise data (50k/100k) across epochs on Open LLM Leaderboard. For comparison, we present the results of PoFT models trained solely with the original SFT data.}
    \label{time_step_neg}
\end{figure}

We conduct experiments on the Mistral-7B model with the OpenHermes datasets as well as the hand-crafted noise data described in section \ref{data_filtering}. Figure \ref{time_step_neg} depicts the performance of bi-PoFT models with different proportions of noise data, compared to the PoFT model trained solely with the original data. 
It can be observed that although the bi-PoFT model under-performs the PoFT model at the initial stages, it exhibits enhanced stability as the training epoch increases. Notably, an increase in noise data proportion correlates with higher peak performance in bi-PoFT models.
These results illustrate the robustness and potential of bi-PoFT models when trained with noise data, surpassing the performance of the PoFT models trained solely on original data.
The disparity could be expanded as an appropriate proportion of noise data involved in the training process, which will be addressed in future work.

\section{Ethics Statement}
This research exclusively employs methods and technologies within the field of Natural Language Processing (NLP). Throughout our experimentation, we strictly adhered to ethical guidelines and rules to ensure that no potential risks or unexpected consequences were caused. The data used in this research does not contain sensitive or offensive content. We aim to contribute positively to the NLP community and advance the technology.

\bibliography{aaai25}

\end{document}